\documentclass[runningheads]{llncs}
\usepackage{graphicx}
\usepackage{amsmath}
\usepackage{xcolor}
\usepackage{hyperref}

\begin{document}
\title{Concept Induction using LLMs: A user experiment for assessment}
%
%
%
\author{
Adrita Barua\inst{1} \and
Cara Widmer\inst{2} \and
Pascal Hitzler\inst{1}
}

\institute{Kansas State University, Manhattan, KS, USA \and Kairos Research, LLC, Dayton, OH, USA} 
\authorrunning{A. Barua, C. Widmer, P. Hitzler}
%

%
\maketitle              
\begin{abstract}
\sloppy
Explainable Artificial Intelligence (XAI) poses a significant challenge in providing transparent and understandable insights into complex AI models. Traditional post-hoc algorithms, while useful, often struggle to deliver interpretable explanations. Concept-based models offer a promising avenue by incorporating explicit representations of concepts to enhance interpretability. However, existing research on automatic concept discovery methods is often limited by lower-level concepts, costly human annotation requirements, and a restricted domain of background knowledge. In this study, we explore the potential of a Large Language Model (LLM), specifically GPT-4, by leveraging its domain knowledge and common-sense capability to generate high-level concepts that are meaningful as explanations for humans, for a specific setting of image classification. We use minimal textual object information available in the data via prompting to facilitate this process. To evaluate the output, we compare the concepts generated by the LLM with two other methods: concepts generated by humans and the ECII heuristic concept induction system. Since there is no established metric to determine the human understandability of concepts, we conducted a human study to assess the effectiveness of the LLM-generated concepts. Our findings indicate that while human-generated explanations remain superior, concepts derived from GPT-4 are more comprehensible to humans compared to those generated by ECII. 

\end{abstract}
\section{Introduction}
\label{sec:intro}
In recent years, Explainable AI (XAI) has emerged as a critical area of research and development within the field of artificial intelligence \cite{angelov2021explainable}. The importance of XAI lies in its ability to enhance transparency and understanding of AI systems, enabling users and stakeholders to comprehend how AI arrives at its decisions. This transparency is crucial for establishing trust in AI systems that may otherwise appear as black boxes or opaque entities \cite{rudin2019stop}, particularly in domains where AI decisions directly impact human lives. Additionally, XAI serves as a means to debug and enhance network architectures to achieve improved outcomes. 

Currently, significant research in XAI is centered on model-agnostic post-hoc algorithms \cite{arrieta2020explainable}. These algorithms aim to provide understandable insights into how a pre-existing model generates predictions for any given input without compromising the model's accuracy. They address the challenge of making complex AI models, such as deep neural networks, more interpretable by humans. Techniques like gradient-based saliency maps \cite{simonyan2014deep} and perturbation-based methods \cite{ivanovs2021perturbation} have been used extensively to identify important features within AI models. Gradient-based methods highlight which parts of the input data influence the model's predictions the most, whereas perturbation-based methods involve modifying the input data to observe changes in the model's output. These approaches employ visual explanation maps to understand the decisions made by a deep learning (DL) model. However, due to challenges in visualizing these explanation maps and considering their susceptibility to adversarial attacks, there has been a shift away from these models \cite{das2020opportunities}.  

One promising approach within XAI is the use of concept-based models for generating explanations \cite{kim2018interpretability,oikarinen2022clip}. Concept-based models incorporate explicit representations of concepts or knowledge units, making them inherently more interpretable than traditional black-box models. By leveraging these explicit concepts, XAI techniques can provide explanations that align with human intuition and reasoning. However, generating meaningful concepts from input data remains a challenge, as it requires context-specific explanations that can accurately depict the model's behavior while remaining understandable to humans.

In previous work \cite{DBLP:conf/nesy/SarkerXDRH17}, a post-hoc explainability strategy was proposed using concept induction \cite{DBLP:journals/ml/LehmannH10,lehmann2014perspectives}, and in \cite{widmer2023towards} it was investigated how well explanations created via concept induction "make sense" for humans. Concept induction involves creating complex Description Logic class descriptions (TBox axioms) based on instance examples (ABox data) using deductive reasoning algorithms over Description Logic knowledge bases. The authors in \cite{widmer2023towards} demonstrated that concept induction can be utilized to explain data differentials in machine learning classifications, although human generated explanations are still better. Their method utilizes the Wikipedia category hierarchy \cite{sarker2020wikipedia} as the background knowledge, and the ECII (Efficient Concept Induction from Individuals) heuristic concept induction system \cite{sarker2019efficient} was used for explanation generation. A survey was conducted through Amazon Mechanical Turk to assess how meaningful the generated explanations are to humans.

Expanding upon this framework, our goal is to explore the feasibility of replacing the ECII model with a Large Language Model (LLM) to produce explanations that remain meaningful and coherent. The objective is to identify "good" concepts that make sense to humans and can later be validated by mapping them with a Deep Neural Network (DNN) to accurately describe what neurons perceive. We utilized the GPT-4 \cite{achiam2023GPT} model to generate meaningful explanations for a specific scene classification task, which was done using a logistic regression algorithm that classified images into scene categories based on semantic tags of objects present in each image. The explanations are generated using Prompt Engineering \cite{ekin2023prompt} via the OpenAI API. Unlike logical-deduction-based systems such as ECII, which are limited by background knowledge, an LLM like GPT-4 can leverage its common-sense reasoning capability and vast domain knowledge to produce more comprehensive concepts. In the mentioned \cite{widmer2023towards}, the quality of explanations generated by concept induction was assessed and found to be more meaningful than semi-random explanations but less accurate than human-generated (gold standard) ones. Our objective is to evaluate the extent to which explanations generated by LLMs align with human-generated explanations and potentially surpass the concept induction system in terms of accuracy and comprehensibility.

Concept induction is a symbolic reasoning task that can be done using provably correct \cite{DBLP:journals/ml/LehmannH10} or heuristic \cite{sarker2019efficient} deduction algorithms over description logic knowledge bases. In this paper, we are attempting to make use of pre-trained LLMs to produce results that are comparable to or even better than those obtained from a concept induction system. In other words, we are making use of an LLM to do better than a symbolic-reasoning-based algorithm, at least in a specific setting. As such, our work contributes to research on neurosymbolic artificial intelligence.

The rest of the paper is structured as follows. In Section \ref{sec:related_work} we review related literature. In Section \ref{sec:approach}, we give a detailed outline of our approach. In Sections \ref{sec:eval} and \ref{sec:results}, we discuss the evaluation method and the resulting outcomes. In Section \ref{sec:conclusion}, we discuss future work and conclude.

\section{Related work}
\label{sec:related_work}
With the increasing need for explainable AI systems, various methods have been proposed to achieve explainability, each with its strengths and limitations. Interpretable models such as decision trees \cite{song2015decision} and linear regression \cite{hope2020linear} inherently offer transparency by design, enabling straightforward explanations based on the model structure. However, these models may lack the complexity and performance of advanced techniques like deep neural networks. With growing demands for more explainable machine learning (ML) \cite{goodman2017european}, there is a rising need for post-hoc methods that can be applied without retraining or modifying the network. 

Post-hoc explanation techniques like LIME \cite{ribeiro2016should} and SHAP \cite{lundberg2017unified} provide ways to explain complex black-box models by approximating local behavior. They generate explanations at the instance level, offering insights into model predictions for specific input features or data points. Another local explanation method uses gradient-based saliency maps \cite{adebayo2018sanity,simonyan2014deep} to highlight the importance of each pixel in an image for the output result. However, these methods can lack trustworthiness and exhibit random biases \cite{ghorbani2019interpretation,alvarez2018robustness}, as explanations are often valid only for a specific data point that may vary significantly across datasets.  

Methods like TCAV \cite{kim2018interpretability} focus on global explanations by utilizing high-level concepts to estimate their importance for predictions, requiring human-provided concepts. Alternatively, ACE \cite{ghorbani2019towards} leverages image segmentation and clustering to curate automated concepts that may lead to some information loss. Other approaches such as Concept Bottleneck Models (CBM) \cite{koh2020concept} and Post-hoc CBM \cite{yuksekgonul2022post} map DNN models to human-understandable concepts but often rely on hand-picked concepts, highlighting the need for automated methods to generate higher-level concepts.  

One approach involves using background or domain knowledge to generate human-understandable explanations via concept induction \cite{widmer2023towards,confalonieri2021using}. However, these methods are constrained by their background knowledge and fail to capture common-sense interpretations evident to humans during explanation generation. Leveraging LLMs has the potential to bridge this gap by automating concept generation, utilizing minimal text-based object information. Another study \cite{oikarinen2023label} employing a similar approach utilizes GPT-3 with a few-shot method to produce automated concepts. Although aiming to reduce human involvement, this method requires a filtering process to refine initial concepts, based on numerical analysis. 

In the field of XAI, explanations cater to end-users or system developers, but in all cases, it remains crucial that explanations make sense to humans. Our study explores the potential of LLMs in generating explanations through human evaluation, aiming to bridge the gap between complex AI systems and human understanding.

\section{Approach}
\label{sec:approach}

Our approach and evaluation setting is essentially the same as in \cite{widmer2023towards}, however instead of their comparison of explanations generated by (1) humans, (2) concept induction, and (3) a semi-random process, we compare (1) human, (2) concept induction, and (3) GPT-4 prompting. We went into the study with the hypothesis that explanations produced by GPT-4 would outperform those produced by concept induction in terms of "meaningfulness to humans," but that they would still not be as good as the human-generated gold standard.

\subsection{Input Dataset}
\label{sec:input_data}
As in \cite{widmer2023towards}, we used the object tags associated with images from the ADE20K dataset \cite{zhou2017scene,zhou2019semantic} as input, in this case for the GPT-4 model via the OpenAI API. This dataset contains approximately 20,000 human-curated images annotated with scene categories and object tags present in the images. We used a selection of 45 image set pairs. Each image set pair consists of two groups of natural images representing distinct scene categories (A and B), with a total of 90 scene categories across all sets. Each set within a pair consisted of eight images selected at random from a particular category. 

These image set pairs were curated in the previous study \cite{widmer2023towards}, and we adopted the same set of pairs to maintain consistency. Although the object tags in the dataset indicate not only the presence of an object but also details such as the number of objects and occlusions, we focused solely on the object labels for our analysis, disregarding additional annotations. 

To generate explanations from the GPT-4 model, we fed the object tags of the images into the model using prompts. Our objective was to describe what distinguished Category A from Category B in each image set pair, where each image set belongs to a specific scene category. These descriptions were defined as "concepts," and for each image set, we produced a list of seven concepts. We tried to come up with concepts that encompass tangible objects depicted in the images (e.g., tree or bench) or general categories that align with the theme of the images (e.g., park or garden).  

To prompt the GPT-4 model effectively, we experimented with different prompting techniques to obtain the most reasonable concepts. Our approach involved using a straightforward technique that leveraged only the object labels from each image set category. We instructed the GPT-4 model to differentiate between the two categories based on their object tags. Object tags, as the name suggests, could be anything physically present in the images. For example, the object tags coming from category A in Figure \ref{fig1} include object labels such as stands, food, wall, tomatoes, bag, register, weighing machine, shopping carts, person, etc.

Similarly, the ECII system also used the same object tags to generate concepts. For the ECII model, all object tags from the images are automatically mapped to classes in the Wikipedia class hierarchy using the Levenshtein string similarity metric \cite{levenshtein1975minimal} with an edit distance of 0. The algorithm then assessed the images based on their object tags and returned a rating of how well concepts matched images in Category A but not Category B. ECII explanations were then created by taking the seven highest-rated unique concepts. This alignment allowed us to compare the concepts generated by our approach with those produced by the ECII system.

The process and the prompt used for interacting with the GPT-4 model are illustrated in Figure \ref{fig1}.

\subsection{Prompting the model}
\label{sec:prompt}
\begin{figure}[tb]
\includegraphics[width=\textwidth]{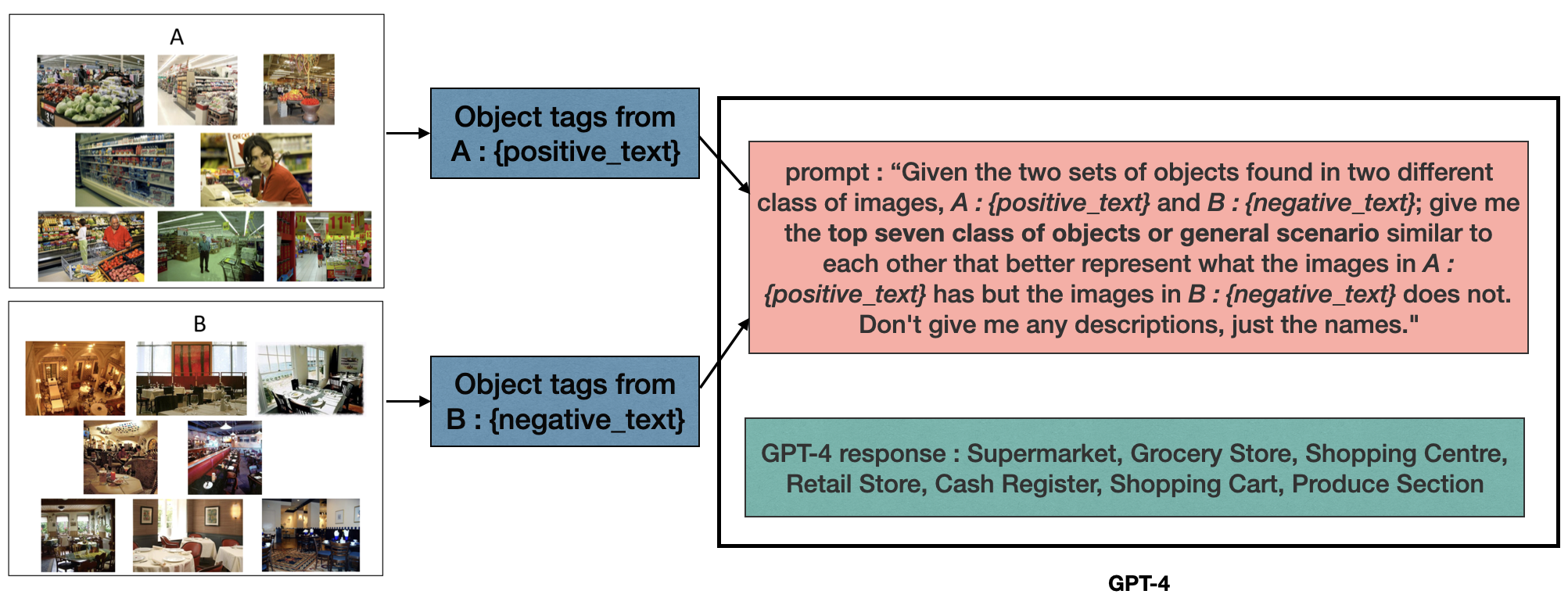}
\caption{Prompting Method: The GPT-4 model was prompted using the exact prompt mentioned in the image. Here, the positive and negative text indicates the object tags present in the images. The resulting set of seven concepts is mentioned in the GPT-4 response.} \label{fig1}
\end{figure}

We used the latest version of the GPT-4 model for our prompt. We utilized zero-shot prompting with specific parameters, setting the temperature to 0.5 and top\_p to 1. The temperature parameter in GPT-4 controls the level of creativity or randomness in the generated text. When predicting the next token from a vocabulary of size \(N\), the model uses a softmax distribution of the form \(\text{softmax}(x_i / T)\) for \(i = 1, \ldots, N\), where \(T\) is the temperature. This distribution assigns probabilities to each token ($x_i$) in the vocabulary, influencing the likelihood of selecting each word. Lowering the temperature favors words with higher probabilities, leading to more predictable and less creative responses when the model randomly samples the next word. Top\_p sampling is an alternative to temperature sampling. It limits the consideration from all possible tokens to a subset of tokens (the nucleus) whose cumulative probability mass reaches a specified threshold (top\_p). OpenAI recommends adjusting one of these parameters but not both simultaneously for optimal control over text generation. In our prompts, we set the model's temperature to a lower value (0.5) to ensure more consistent and reproducible answers across different sets. Here, we didn't set the temperature to 0 as we wanted to see some creative responses from the GPT-4 model in tasks where the image set categories (e.g., Category A and B) contain similar objects, to test if the model can distinguish them using human-like intuitive behavior. In figure \ref{fig1}, we can see that all the object tags coming from sets A and B are given in the prompt, and it was asked to distinguish between them. Here as it becomes a long prompt with all the object tags for both categories, we mention them twice in our prompt, once at the beginning and once at the end, which seems to be helpful for the GPT model to produce better results and remember the object tags. In our prompts, we aimed to generate generic concepts or object classes that mimic the ontology classes positioned somewhere in the middle of the hierarchy used by ECII. These intermediate classes are designed to capture a broader range of specific child classes, providing a bridge between more general concepts and highly specific subclasses within the ontology structure. It is asked to provide the top seven concepts based on the instruction. We generate a list of seven concepts for each set following this method.

\section{Evaluation}
\label{sec:eval}
To evaluate the concepts generated from GPT-4 we ran a study through Amazon Mechanical Turk using the Cloud Research platform. Our goal was to assess the quality of LLM explanations (i.e., GPT-4 explanations) compared to both human-generated ("gold standard") explanations and ECII explanations. The study protocol was reviewed and approved by the Institutional Review Board (IRB) at Kansas State University and was deemed exempt under the criteria outlined in the Federal Policy for the Protection of Human Subjects, 45 CFR §104(d), category: Exempt Category 2 Subsection ii \ref{sec:appendices}.

We recruited 300 participants through Mechanical Turk, compensating each participant with \$5 for completing the task, which was estimated to take approximately 40 minutes (equivalent to \$7.50 per hour based on the minimum legal wage in the USA). Based on the previous study \cite{widmer2023towards}, we aimed for a sample size of at least 89 unique participant judgments per trial to estimate the parameters (medium effect size of f2 = 0.15 and 95\% power) of the Bradley-Terry model \cite{bradley1952rank}, which is used to evaluate the survey results. This required collecting data from 300 participants, resulting in a total of 100 observations per trial after accounting for potential exclusions.  

Across all questions, each participant encountered three types of explanations, although only two explanation types were compared in any given question. Each participant was asked to choose the more accurate explanation using a two-alternative forced choice design. For each pair of image sets, participants answered three questions comparing (1) Human versus ECII explanation; (2) Human versus LLM (GPT-4) explanation; and (3) LLM versus ECII explanation. For each pair of image sets (A and B), a given participant completed all three comparisons.

The 45 pairs of image sets in this study resulted in a total of 135 unique target questions. Participants were randomly assigned to 15 image sets (45 questions in total), ensuring that image sets were counterbalanced across participants to receive an equal number of responses.

For all image sets, ECII explanations and Human "gold standard" explanations were created in a previous study \cite{widmer2023towards}. In this work, we generated LLM (GPT-4) explanations following the method described in Section \ref{sec:approach}. To form the ECII explanations, the object tags of the images were provided to the ECII algorithm, then the seven highest-rated unique concepts were selected based on the ranking of the F1 score. Human "gold standard" explanations were crafted by presenting image sets (without object or scene category tags) to three human raters, selecting concepts unanimously mentioned by all three, then by two raters, and finally filling in randomly selected concepts until seven unique concepts were reached.

In addition to the 45 image sets, five "catch trial" image sets were used to verify participant attention. These catch-trial image sets included two types of explanations: human explanations generated similarly to other human gold standard explanations, and explanations consisting of completely random concepts generated from a word generator to serve as obviously inaccurate explanations. 

\begin{figure}[tb]
\includegraphics[width=\textwidth]{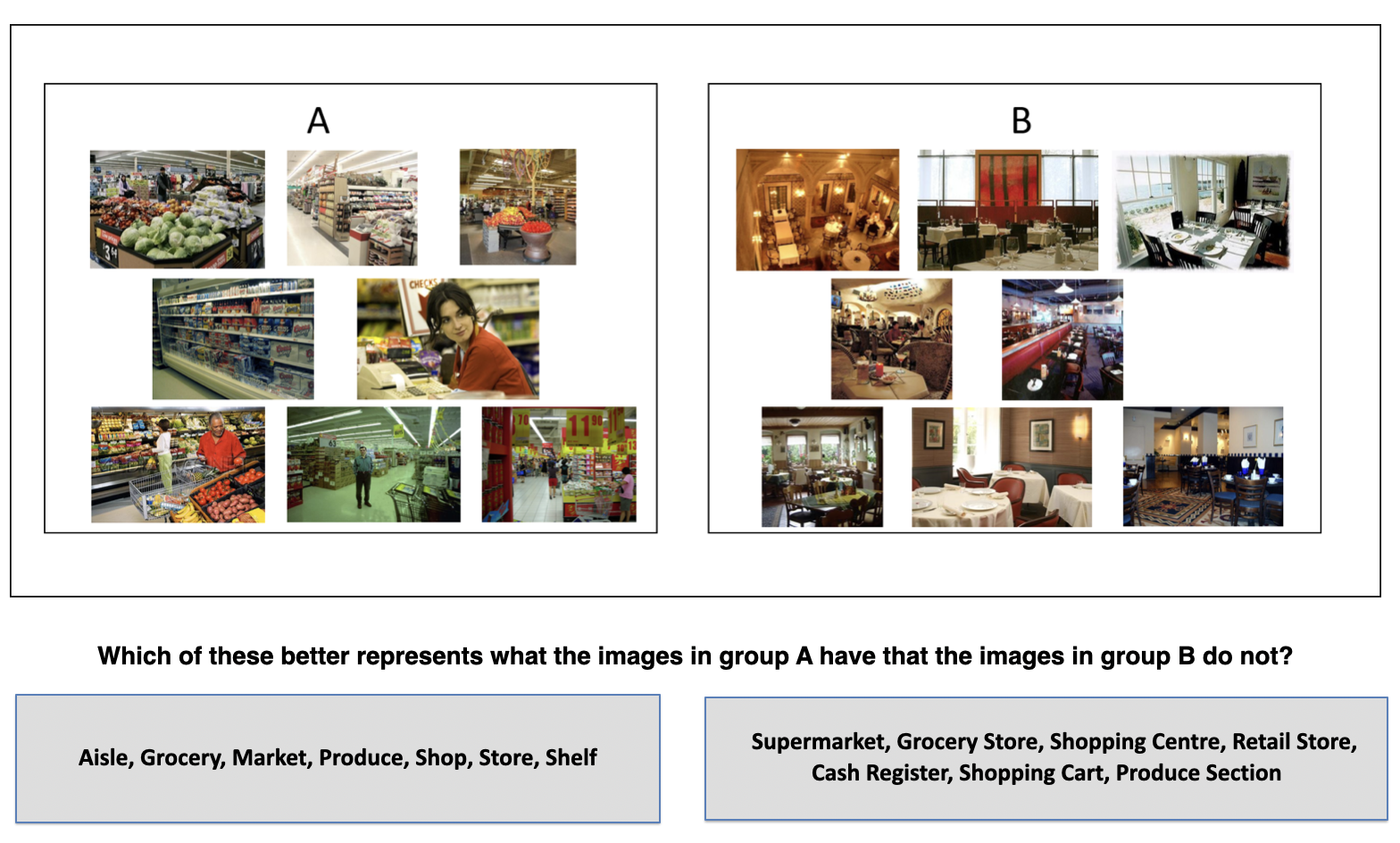}
\caption{ Survey interface, with human explanation presented on the left and LLM explanation on the right.} \label{survey}
\end{figure}

After providing consent, participants received brief training on the task, including instructions on how concepts and explanations were defined in the study. They then began answering questions, with the 50 questions (45 assigned targets and 5 catch trials) presented in random order. Figure \ref{survey} illustrates the stimuli presentation and response options shown to participants.

\section{Results}
\label{sec:results}
Prior to analysis, participant responses to catch trials were evaluated, and participants who failed more than one catch trial were excluded from further analysis. Among the 300 participants, 253 did not fail any catch trials, while 22 participants failed exactly one trial, and 35 participants failed more than one trial. The 35 participants who failed multiple trials were excluded from all subsequent analyses, resulting in a total of 265 participants included in the analyses.

\begin{figure}[tb]
\centering
\includegraphics[width=.8\textwidth]{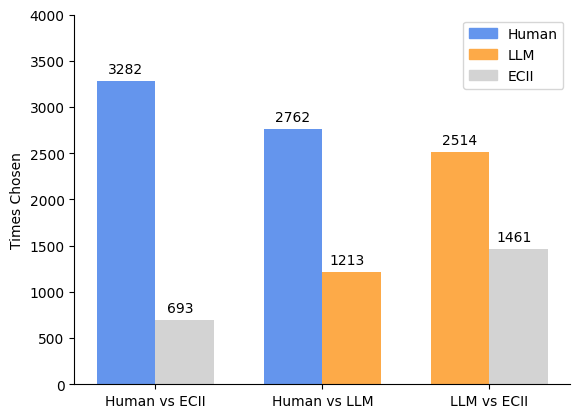}
\caption{ Number of times participants chose different explanation types.} \label{fig2}
\end{figure}

Across all image sets, human explanations were overwhelmingly preferred over ECII explanations (chosen 3282 times versus 693 times; 83\% preference) and over LLM (GPT-4) explanations (chosen 2762 times versus 1213 times; 69\% preference). Additionally, LLM explanations were preferred over ECII explanations (chosen 2514 times compared to 1461 times; 63\% preference). See Figure~\ref{fig2}.

Participants' pairwise judgments were utilized in a Bradley-Terry analysis \cite{firth2012bradley} to derive "ability scores" for each type of explanation, reflecting the extent to which each explanation type was preferred by participants.  The Bradley-Terry model uses data where entities are compared pairwise, and the outcome (win/loss, preference ranking, etc.) is observed. From these comparisons, the model estimates the abilities \(\theta_i\) such that the observed outcomes are statistically likely. The estimation process typically involves fitting the model to the pairwise comparison data to find the best-fitting values of \(\theta_i\) for each entity. These estimates reflect the latent abilities or strengths of the entities relative to each other. Ability scores were calculated for each of the 45 image set pairs based on the pairwise comparison data (win/loss) for each type of explanation. The analysis of these ability scores demonstrated that human explanations had the highest scores (M = 1.77, SD = 0.978), followed by LLM explanations (M = 0.724, SD = 1.16), with a significant overall difference (F(2) = 46.28, $p < 0.001$, $\eta^2 = 0.41$). Here, ECII explanations served as the reference point and were set to 0, with the ability scores for human and LLM explanations indicating their preference over ECII explanations. 

A post hoc analysis using Tukey’s Honestly Significant Difference (HSD) test \cite{abdi2010tukey} was conducted to determine which specific group means are significantly different from each other. When comparing multiple group means, the Tukey post hoc test is preferred over multiple t-tests \cite{kim2015t} because it adjusts for multiple comparisons, controlling the overall Type I error rate \cite{nanda2021multiple}. Conducting multiple t-tests increases the risk of false positives, while the Tukey test maintains the integrity of statistical conclusions by adjusting the significance levels appropriately. This test confirmed significant differences in ability scores between human vs. ECII explanations and human vs. LLM explanations (both $p < 0.0001$), as well as between LLM vs. ECII explanations ($p = 0.0004$) (Table \ref{tab2}). These low p-values indicate that the observed differences in ability scores are highly significant and unlikely to have occurred by random chance alone.

The individual ability scores for human and LLM explanations for each image set pair are detailed in Table \ref{tab1}. 

\begin{table}
\caption{Ability Scores and Number of Wins for Human (H), ECII (E), and LLM (L) explanations. ECII explanations were set as the reference point in the Bradley-Terry analysis and so their ability scores were always equal to 0, and thus are not displayed here.}\label{tab1}
\resizebox{\columnwidth}{!}{%
\begin{tabular}{l|cc|ccc}
\hline
\multicolumn{1}{l|}{Image Set}         & \multicolumn{1}{l}{H.Ability} & \multicolumn{1}{l|}{LLM.Ability} & \multicolumn{1}{l}{HvE Wins} & \multicolumn{1}{l}{HvL Wins} & \multicolumn{1}{l}{LvE Wins} \\ \hline
Set 1: Bedroom v Park                   & 1.47                           & -1.05                            & 72-12                         & 74-10                         & 18-66                         \\
Set 2: Living Room v Parking Lot        & 2.64                           & 2.76                             & 84-3                          & 38-49                         & 79-8                          \\
Set 3: Office v Playground              & 1.12                           & 0.34                             & 74-15                         & 54-35                         & 45-44                         \\
Set 4: Airport v Amusement Park         & 1.93                           & 0.77                             & 77-13                         & 70-20                         & 63-27                         \\
Set 5: Bathroom v Art Studio            & 1.05                           & 1.47                             & 67-20                         & 32-55                         & 68-19                         \\ \hline
Set 6: Beauty Salon v Forest Path       & 0.72                           & -0.86                            & 63-25                         & 69-19                         & 22-66                         \\
Set 7: Bookstore v Child Room           & 1.72                           & 1.79                             & 76-15                         & 45-46                         & 79-12                         \\
Set 8: Hotel Room v Cockpit             & 0.65                           & -1.68                            & 62-28                         & 79-11                         & 11-79                         \\
Set 9: Shoe Store v Alcove              & 0.79                           & 1.52                             & 63-24                         & 25-62                         & 68-19                         \\
Set 10: Alley v Wet Bar                 & 2.74                           & 1.85                             & 85-6                          & 65-26                         & 79-12                         \\ \hline
Set 11: Closet v Construction Site      & 1.98                           & 1.14                             & 77-8                          & 57-28                         & 62-23                         \\
Set 12: Gazebo v Bowling Alley          & 2.64                           & -1.03                            & 85-2                          & 81-6                          & 19-68                         \\
Set 13: Garage v Hallway                & 0.42                           & -0.09                            & 49-39                         & 59-29                         & 46-42                         \\
Set 14: Laundromat v Pantry             & 1.86                           & 1.18                             & 75-14                         & 61-28                         & 70-19                         \\
Set 15: Conference Room v Waterfall      & 2.42                           & -0.45                            & 85-3                          & 79-9                          & 30-58                         \\ \hline
Set 16: Home Office v Bow               & 1.83                           & 1.58                             & 77-13                         & 51-39                         & 75-15                         \\
Set 17: Dining Room v Kitchen           & 0.24                           & 0.33                             & 45-41                         & 44-42                         & 53-33                         \\
Set 18: Fast Food v Office Building     & 2.58                           & 0.24                             & 84-4                          & 78-10                         & 47-41                         \\
Set 19: Jacuzzi v Greenhouse            & 3.08                           & 2.13                             & 88-5                          & 68-25                         & 84-9                          \\
Set 20: Gymnasium v Corridor            & 2.76                           & 1.63                             & 83-6                          & 68-21                         & 75-14                         \\ \hline
Set 21: Bus v Broadleaf Forest          & 2.24                           & -0.59                            & 77-8                          & 80-5                          & 30-55                         \\
Set 22: Casino v Arrival Gate           & 1.77                           & 1.11                             & 73-13                         & 57-29                         & 65-21                         \\
Set 23: Library v Gas Station           & 0.92                           & -1.02                            & 61-31                         & 85-7                          & 29-63                         \\
Set 24: Valley v Yard                   & 2.66                           & 1.17                             & 85-7                          & 76-16                         & 71-21                         \\
Set 25: Mountain v Coast                & 0.45                           & -0.64                            & 50-36                         & 67-19                         & 32-54                         \\ \hline
Set 26: Dinette Vehicle v Farm Field    & 0.88                           & -0.62                            & 69-23                         & 71-21                         & 28-64                         \\
Set 27: Poolroom v Driveway             & -0.72                          & -0.12                            & 30-58                         & 30-58                         & 40-48                         \\
Set 28: Bridge v Auditorium             & 1.95                           & 1.9                              & 80-10                         & 45-45                         & 77-13                         \\
Set 29: Museum v Youth Hostel           & 1.24                           & -1.04                            & 68-20                         & 80-8                          & 23-65                         \\
Set 30: Supermarket v Restaurant        & 2.12                           & 2.97                             & 75-8                          & 24-59                         & 78-5                          \\ \hline
Set 31: Classroom v Archive             & 1.18                           & 0.06                             & 65-18                         & 61-22                         & 41-42                         \\
Set 32: Dentist Office v Ballroom       & 2.94                           & 1.29                             & 85-5                          & 76-14                         & 71-19                         \\
Set 33: Lighthouse v River              & 1.68                           & 1.81                             & 73-14                         & 41-46                         & 75-12                         \\
Set 34: Creek v Basement                & 4.46                           & 2.85                             & 86-4                          & 78-12                         & 88-2                          \\
Set 35: Building Facade v Ocean         & 1.69                           & 0.77                             & 77-16                         & 68-25                         & 65-28                         \\ \hline
Set 36: Courthouse v Parking Garage     & 2.95                           & 1.15                             & 82-7                          & 79-10                         & 70-19                         \\
Set 37: Balcony v Skyscraper            & 3.18                           & 0.8                              & 84-4                          & 81-7                          & 61-27                         \\
Set 38: Game Room v Waiting Room        & 0.68                           & 0.09                             & 63-29                         & 57-35                         & 46-46                         \\
Set 39: Landing Deck v Window Seat      & 2.72                           & 2.15                             & 86-4                          & 56-34                         & 79-11                         \\
Set 40: Bar v Warehouse                 & 1.35                           & 0.47                             & 73-15                         & 59-29                         & 51-37                         \\ \hline
Set 41: Bakery v Apartment Building & 0.99                           & 1.98                             & 63-21                         & 21-63                         & 72-12                         \\
Set 42: Needleleaf Forest v Playroom        & 2.41                           & 1.14                             & 81-8                          & 70-19                         & 68-21                         \\
Set 43: Outdoor Window v Roundabout      & 2.14                           & 0.53                             & 84-8                          & 75-17                         & 56-36                         \\
Set 44: Reception v Golf Course         & 2.16                           & 0.99                             & 76-9                          & 65-20                         & 62-23                         \\
Set 45: Staircase v Plaza               & 1.09                           & 0.04                             & 65-21                         & 63-23                         & 43-43                        
\end{tabular}}
\end{table}

\begin{table}
\caption{p-values of the ability scores among different explanation types from Tukey’s HSD Test}\label{tab2}
\centering
\begin{tabular}{l|l}
Comparison pairs                       &  p-value          \\
\hline
Human\_explanation vs ECII\_explanation & \textless 0.0001 \\
LLM\_explanation vs ECII\_explanation   & =0.0004           \\
Human\_explanation vs LLM\_explanation  & \textless 0.0001 \\ 
\end{tabular}
\end{table}

The source code, input data, and raw result files related to the evaluation tasks (i.e., survey questionnaires, and collected responses) are available online \href{https://github.com/AdritaBarua/Concept-Induction-using-LLMs-a-user-experiment-for-assessment}{here}.

\subsection{Discussion}
\label{sec:discussion}
The analysis of the results presented in Table \ref{tab1} provides evidence supporting our hypothesis that LLM (GPT-4) explanations are more meaningful for humans compared to ECII-generated ones. Human-generated explanations were consistently preferred as the most accurate in describing differences between image categories, followed by LLM explanations, with ECII explanations found as the least accurate. 
The preference for human-generated explanations over LLM explanations is expected given the messy nature of generalized Large Language Models. These models, trained on vast and diverse datasets, can produce responses that lack precision and clarity because of their broad generalization. This can lead to explanations that are sometimes inaccurate or unclear, making human-generated explanations generally more reliable and preferred. Also, there is potential for further refinement in prompting techniques using varied hyper-parameters (e.g., temperature and top-p). However, LLM explanations demonstrated notable explanatory power, suggesting their usability in concept generation.

\begin{figure}[tb]
\includegraphics[width=\textwidth]{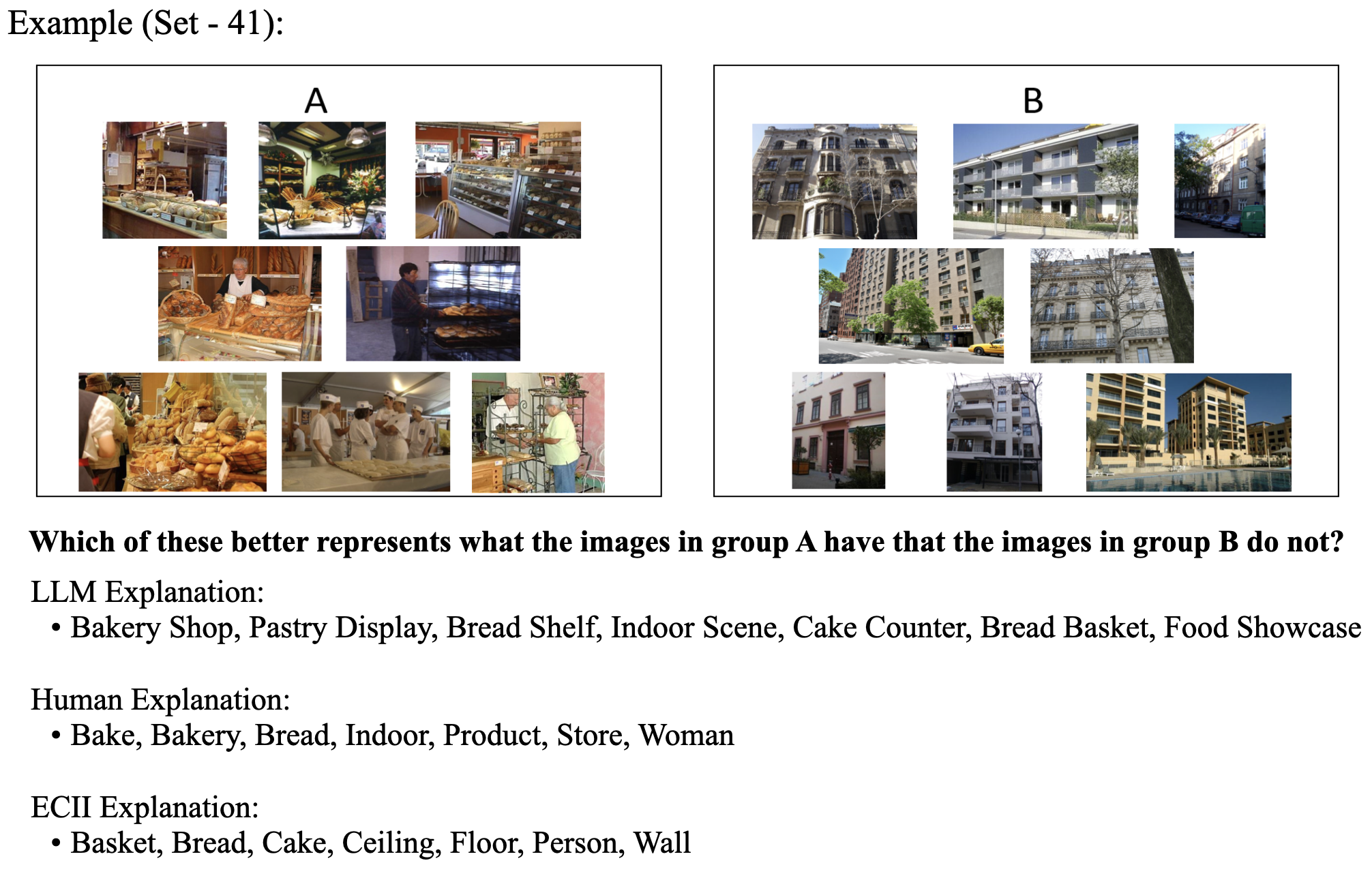}
\caption{ Example of different explanation types for Set 41: Bakery v Apartment Building} \label{set_41}
\end{figure}

\begin{figure}[tb]
\includegraphics[width=\textwidth]{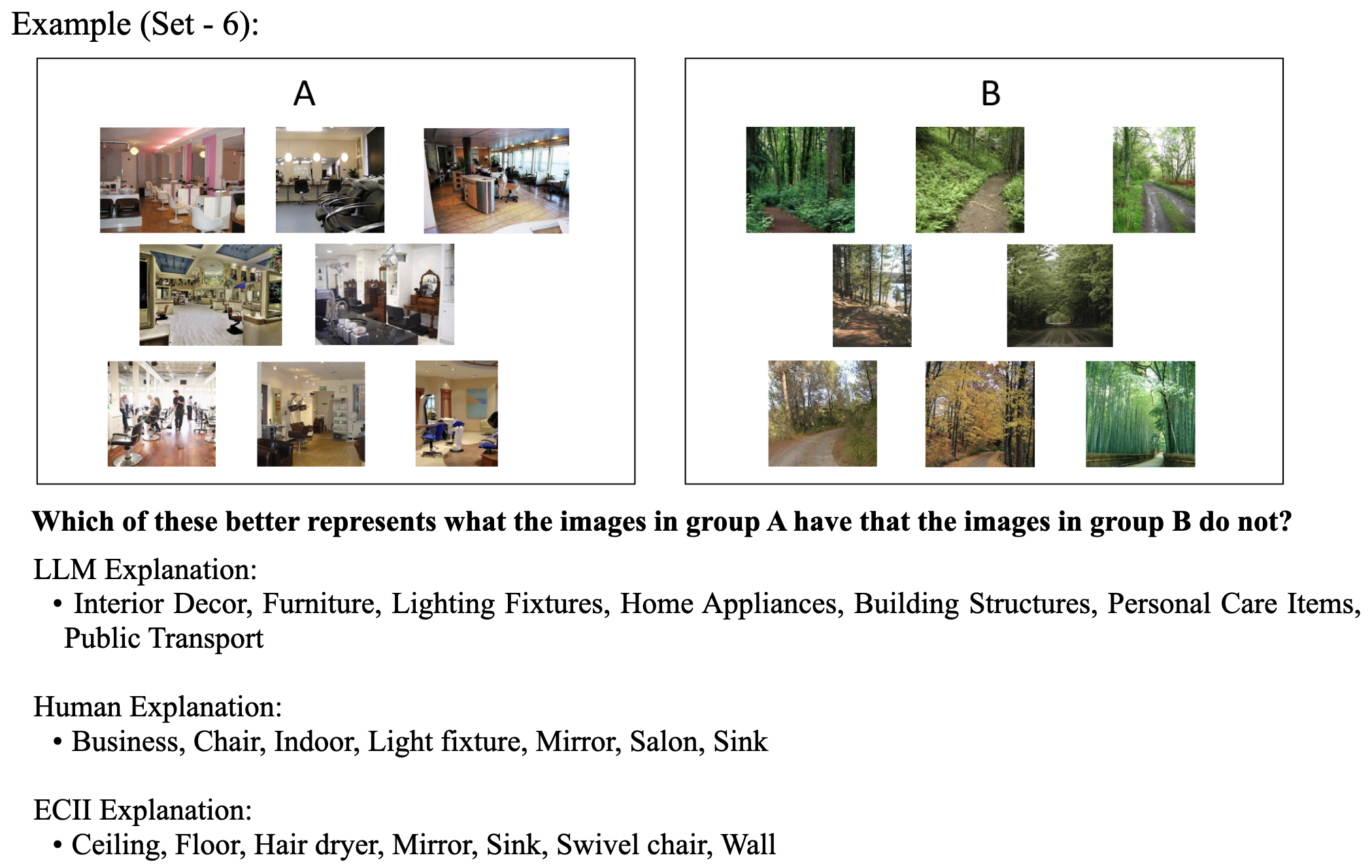}
\caption{ Example of different explanation types for Set 6: Beauty Salon v Forest Path } \label{set_6}
\end{figure}

It is important to note the variability in LLM performance across different image sets. In some cases, LLM explanations were chosen relatively more frequently than in others, with some instances showing LLM explanations being preferred more often than human explanations. Conversely, in other image sets, LLM explanations were chosen less often than ECII explanations. For instance, in Set 41 (see Figure \ref{set_41}), explanations generated by LLM are more comprehensive in identifying images of a bakery, while human-generated explanations also perform adequately. However, the concept "Women" included in the human-generated list is not as relevant for capturing the overall scene depicted in these images.
On the other hand, ECII concepts only identify the object names present in the images and fail to capture the broader category of the scenes (i.e., bakery). In most cases where LLM explanations fall short, they tend to introduce concepts that are unrelated to the images. For example, in Set 6 (see Figure~\ref{set_6}), LLM produced a concept like "Public Transport," which is contextually incorrect. One potential reason for this is the presence of object names (such as streetcar, tram, tramcar, swivel chair, trolley car, armchair) in the input images, which could be erroneously associated with public transport. Based on these examples, it is speculated that when GPT-4 was prompted to generate generic scenarios based on object tags, it attempted to produce seven distinct concepts. Limiting the number of concepts might lead to clearer explanations that are more pertinent. Additionally, running prompts to ask for simple object names rather than generic scenarios akin to ECII-generated explanations could yield different outputs that may prove useful. This suggests there is certainly still room for improvement in LLM explanations, but that on average there is promising evidence that LLMs can produce explanations that successfully describe the differences between two groups of data. Moreover, variability could be introduced by human participants. In our study, human explanations were preferred over ECII 83\% of the time, whereas in the previous study \cite{widmer2023towards} with the same settings, the preference ratio was 87\%.

\section{Conclusions and Future Work}
\label{sec:conclusion}
Based on the findings from our human assessment study of concepts generated by Large Language Models (e.g., GPT-4), it is evident that LLMs hold significant promise in automating concept discovery for complex AI systems. Our study demonstrated that LLMs can produce high-level explanations that are comprehensive and understandable to humans, showcasing their potential to enhance explainability in AI.

The advantage of LLMs over traditional logical concept induction systems lies in their ability to overcome limitations associated with limited background knowledge, algorithmic constraints of heuristic search, and the integration of common sense into explanations. However, we acknowledge the limitations in the current approach of prompting LLMs like GPT-4, particularly in generating irrelevant explanations due to their uncontrolled nature(e.g., hallucination) and reliance on annotated object information in the dataset. These shortcomings can be addressed through improved prompts with varied hyper-parameters and the integration of vision-based models for automatic object identification. Additionally, we thank a reviewer for suggesting the use of a larger sample of images to strengthen the analysis. Further analysis of the similarity between ECII- and LLM-induced concepts, for example, using a similarity model, would also be interesting to see. We want to explore further research on fine-tuning open-source LLMs with a symbolic approach using description logic and few-shot training can create a more controlled environment for meaningful concept generation, aligning closely with a neurosymbolic approach to Explainable AI.

While it is crucial to ensure that LLM-generated concepts are meaningful to human users, further development is needed to assess their mapping with network activations and performance under controlled conditions. We aim to advance the development of an automated system that harnesses LLMs to offer descriptive explanations that are both human-understandable and verifiable through deep neural network activations.

This study serves as a testament to the efficient utilization of LLMs in the domain of Concept Induction and lays the groundwork for future research in leveraging these models to enhance the explainability of AI systems.

\paragraph{Acknowledgement}The authors acknowledge partial funding under the National Science Foundation grant 2333782 "Proto-OKN Theme 1: Safe Agricultural Products and Water Graph (SAWGraph): An OKN to Monitor and Trace PFAS and Other Contaminants in the Nation's Food and Water Systems."

%
%
%
%
\bibliographystyle{splncs04}
\bibliography{ref}

\newpage
\section{Appendices}
\label{sec:appendices}
The IRB approval for the study is attached here. 
\begin{figure*}[h]
\includegraphics[width=0.9\textwidth]{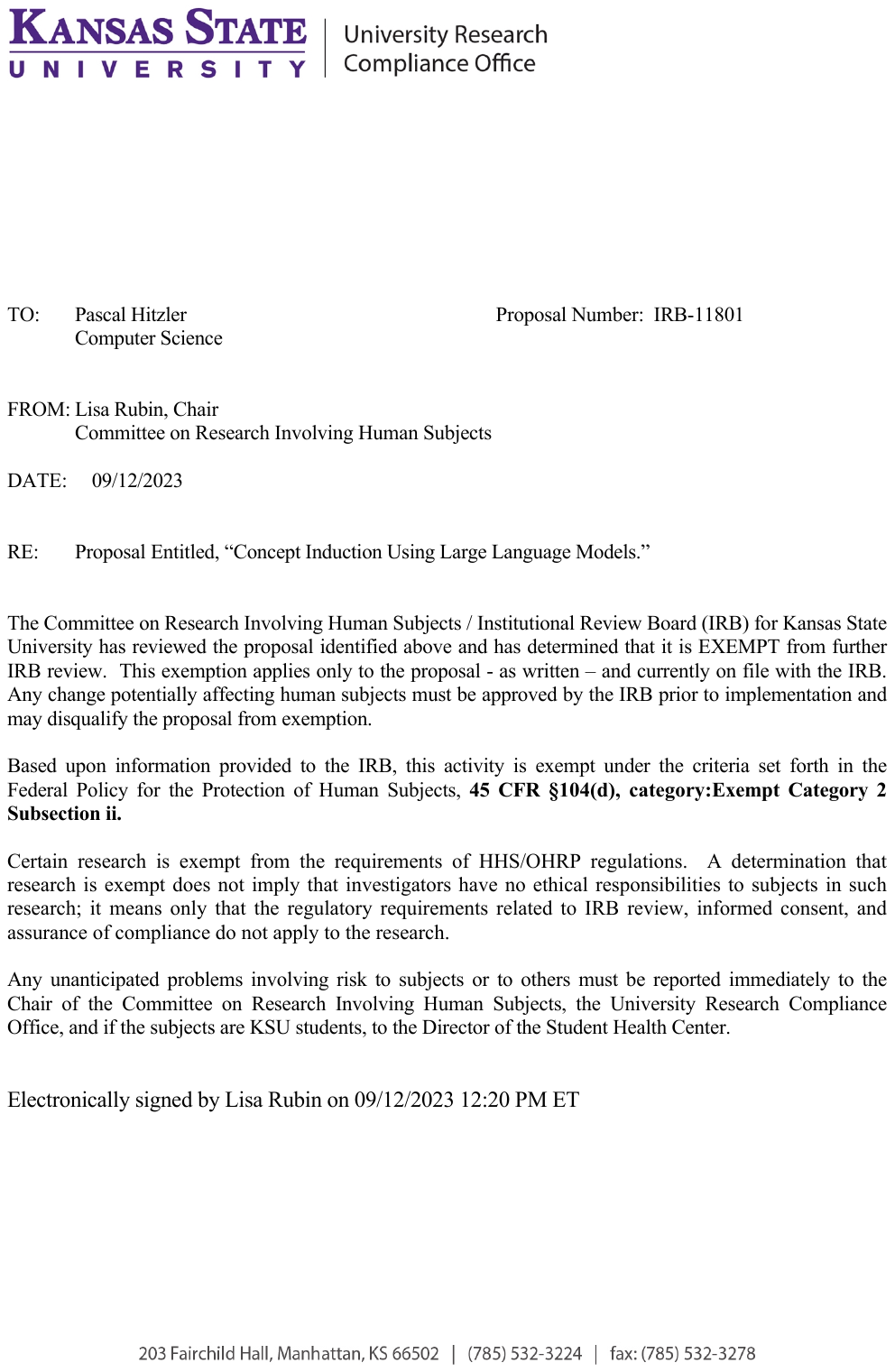}
\label{IRB}
\end{figure*}

\end{document}